\documentclass[twocolumn]{article}
\usepackage{arxiv}
\ifpdf \usepackage[pdftex]{graphicx} \pdfcompresslevel=9
\else \usepackage[dvips]{graphicx} \fi
\usepackage[utf8]{inputenc}
\usepackage{lipsum}
\usepackage{hyperref}
\usepackage{microtype}
\usepackage{subfigure}
\usepackage{booktabs}
\usepackage{amsfonts}       
\usepackage{graphicx}
\usepackage{amsmath}
\usepackage{comment}

\usepackage{amssymb} 
\usepackage{multirow}
\usepackage{arydshln}

\usepackage{t1enc}
\usepackage{cite}

\title{STaRFlow: A SpatioTemporal Recurrent Cell for Lightweight Multi-Frame Optical Flow Estimation}

\author{Pierre Godet$^1$, Alexandre Boulch$^2$, Aurélien Plyer$^1$, and Guy Le Besnerais$^1$\\
$^1$ DTIS, ONERA, Université Paris-Saclay, FR-91123 Palaiseau, France\\
Email: \texttt{\{pierre.godet, aurelien.plyer, guy.le\_besnerais\}@onera.fr}\\
$^2$ valeo.ai, Paris, France\\
Email: \texttt{alexandre.boulch@valeo.com}
}

\begin{document}

\maketitle

\begin{abstract}

We present a new lightweight CNN-based algorithm for multi-frame optical flow estimation.
Our solution introduces a double recurrence over spatial scale and time through repeated use of a generic "STaR" (SpatioTemporal Recurrent) cell. It includes (i) a temporal recurrence based on conveying learned features rather than optical flow estimates; (ii) an occlusion detection process which is coupled with optical flow estimation and therefore uses a very limited number of extra parameters. The resulting STaRFlow algorithm gives state-of-the-art performances on MPI Sintel and Kitti2015 and involves significantly less parameters than all other methods with comparable results.

\end{abstract}

\section{Introduction}
Optical Flow (OF) is the apparent displacement of objects between two frames of a video sequence. It expresses the direction and the magnitude of the motion of each object at pixel level. The OF is
a key component for several computer vision tasks, such as action recognition~\cite{simonyan2014two}, autonomous navigation~\cite{janai2017computer}, tracking~\cite{chen2011tracking}, or image registration for multi-view applications like video inpainting~\cite{xu2019deep}, super-resolution~\cite{zhao2002super,mitzel2009video,nasrollahi2014super} or structure from motion~\cite{ummenhofer2017demon}.
OF estimation must be fast, accurate even at subpixel level for some applications like super-resolution, and reliable even at sharp motion boundaries despite occlusion effects.
Particularly, it must deal with challenging contexts such as fast motions, motion blur, illumination effects, uniformly colored objects, etc.

Starting from the seminal work of Horn and Schunck~\cite{horn1981determining}, OF estimation has been the subject of numerous works.
Recently, a breakthrough came with deep neural networks.
Convolutional neural network-based (CNNs) methods~\cite{dosovitskiy2015flownet,mayer2016large,ilg2017flownet2,sun2018pwc} reached the state of the art on mostly all large OF estimation benchmarks, e.g.,  MPI Sintel \cite{butler2012sintel} and Kitti \cite{Menze2015CVPR}, while running much faster than previous variational methods.

In order to increase the efficiency and the robustness of these methods, the focus has then been put on occlusion detection~\cite{ilg2018occlusions,neoral2018continual,hur2019iterative}, temporal dependency~\cite{neoral2018continual} or memory efficiency~\cite{hui18liteflownet,hur2019iterative}.
Building on these concerns, our work follows two main orientations.
First, when processing a video sequence, most object motions are continuous across frame pairs. Thus, most of the uncertainties arising from two-frame OF estimation can be solved using a number of frames greater than two. This calls for a multi-frame estimation process able to exploit temporal redundancy of the OF.
Second, we believe that related operations can be performed by identical models with shared weights. We apply this principle to temporal recurrence, as in \cite{neoral2018continual}, to scale recurrence, as in \cite{hur2019iterative}, but also to occlusion detection, which is strongly correlated with OF estimation.
Based on these considerations, we propose a "doubly recurrent" network over spatial scales and time instants. It takes explicitly into account the information from previous frames and the redundancy of the estimation at each network scale within a unique processing cell, denoted \textit{STaR} cell, for \textit{S}patio\textit{T}empor\textit{a}l \textit{R}ecurrent cell.
Given information from the past and from a lower scale, the STaR cell outputs the OF and occlusion map at current image scale and time instant. This cell is repeatedly invoked over scales in a coarse-to-fine scheme and over sets of $N$ successive frames, leading to the STaRFlow model.
Thanks to this doubly recurrent structure, and by sharing the weights between processes dedicated to flow estimation and to occlusion detection, we obtain a lightweight model: STaRFlow is indeed slightly lighter than LiteFlowNet~\cite{hui18liteflownet}, while producing jointly multi-frame OF estimation and occlusion detection.

\begin{figure*}[ht]
\begin{center}
    \includegraphics[width=1\linewidth]{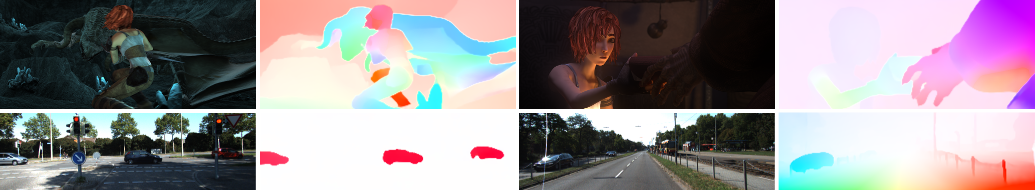}
\end{center}
\caption{Qualitative results of the proposed STaRFlow model, on MPI Sintel final pass (top line) and KITTI 2015 (bottom line) test sets. StaRFlow allows accurate motion estimation on partially occluded objects (right knee of character in upper leftmost example) and on thin objects (fingers and posts in the rightmost examples).}
\label{fig:test_final}
\end{figure*}

Let us now outline the organization of the paper while listing our main contributions.
We first discuss related work in Section~\ref{sec:related}, then Section~\ref{sec:method} is devoted to the description of our main contribution, the STaRFlow model for multi-frame OF estimation. Experiments are presented in Section~\ref{sec:experiments}, with results on MPI Sintel \cite{butler2012sintel} and Kitti \cite{Menze2015CVPR}: examples of results of STaRFlow on these two datasets are presented in Figure~\ref{fig:test_final}. We conduct in particular an ablation study that addresses three important subjects: temporal recurrence, occlusions and scale recurrence. First, as regards temporal recurrence, we show that passing learned features between instants compares favourably to passing previously estimated OF as in ContinualFlow~\cite{neoral2018continual}. Our approach also makes a higher benefit from larger number of frames than~\cite{neoral2018continual}. Secondly, our occlusion handling appears as efficient as previously published approaches, but is much simpler and involves a significantly lower number of extra parameters. Thirdly, the study of scale recurrence highlights the compactness of our model. Finally, concluding remarks and perspectives are given in Section~\ref{sec:conclusion}.

\section{Related Work}\label{sec:related}

\subsection{Optical Flow (OF) Estimation With CNN}
Dosovitskiy \textit{et al.} \cite{dosovitskiy2015flownet} were the first to publish a deep learning approach for OF estimation. They proposed a synthetic training dataset, FlyingChairs, and two CNN architectures FlowNetS and FlowNetC. They have shown fairly good results, though not state-of-the-art, on benchmarks data which are very different from their simple 2D synthetic training dataset.
By using a more complex training dataset, FlyingThings3D \cite{mayer2016large}, and a bigger architecture involving several FlowNet blocks, Ilg \textit{et al.} \cite{ilg2017flownet2} proposed the first state-of-the-art CNN-based method for OF estimation. Moreover, their learning strategy (FlyingChairs $\rightarrow$ FlyingThings) was then used by several supervised learning approaches.

Some of the works that followed~\cite{ranjan2017optical,hui18liteflownet,sun2018pwc} sought to leverage well-known classical practices in OF estimation, like warping-based multi-scale estimation, within a deep learning framework, leading to
state-of-the-art algorithms \cite{hui18liteflownet,sun2018pwc}. In particular, PWC-Net of \cite{sun2018pwc} has then been used as a baseline for several top-performing methods \cite{neoral2018continual,ren2019fusion,liu2019selflow,hur2019iterative,bar2020scopeflow}.
Very recently, Hur and Roth \cite{hur2019iterative} got even closer to classical iterative OF estimation processes with an "iterative residual refinement" (IRR) version of PWC-Net. IRR mainly consists in using the same learned parameters for every stage of the decoder, so as to obtain a lighter and better-performing method. We exploit the same idea but extend it to scale \emph{and temporal} iterations in a multi-frame setting.

\subsection{Multi-Frame Optical Flow Estimation}
Exploiting temporal coherence as been proven to improve estimation quality. Wang \textit{et al.} \cite{wang2008estimating} use multiple frames in a Lucas-Kanade \cite{lucas1981iterative} estimation process and show better results when increasing the number of frames, \textit{i.e.} a less noisy estimation and a reduced number of ambiguous matching points. Volz \textit{et al.} \cite{volz2011modeling} also improve their estimate, in particular in untextured regions, by modeling temporal coherence with an adaptive trajectory regularization in a variational method.
Kennedy and Taylor \cite{kennedy2015optical} shown improved results on the MPI Sintel benchmark \cite{butler2012sintel} by using additional frames, more significantly in unmatched regions.

Additional frames are useful to cope with occlusions, as, for instance, pixels visible at time $t$ and occluded at time $t+1$ may have been visible at time $t-1$. Hence, the OF is ill-defined from $t$ to $t+1$ but can be filled in with the estimation at the previous time step. Ren \textit{et al.} \cite{ren2019fusion} propose a multi-frame fusion process to fuse the current OF estimate
with the estimate at the previous time step. Maurer and Bruhn \cite{maurer2018proflow} propose to learn, with a CNN, how to infer the forward flow from the backward flow, and fuse it with the actual estimated forward flow.
Note that in these references, the multi-frame estimation stems from the fusion of two OF estimates provided by classical two-frame processes launched between different frame pairs.
In contrast, in the ContinualFlow model of \cite{neoral2018continual}, a temporal connection is introduced to pass the OF estimate at time $t-1$ to the estimation process at time $t$, making the estimation recurrent in time.
Let us also mention that, in an unsupervised learning framework, \cite{janai2018unsupervised}, \cite{liu2019selflow} and \cite{liu2020arflow} also show improved results, more significantly in occluded areas, by using multiple frames. These methods use 3 frames and estimate jointly the OF from $t$ to $t-1$ and from $t$ to $t+1$.

Our work is closer to \cite{neoral2018continual}, as we propose to use a recurrent temporal connection, but is based on passing learned features from one instant to the next rather than OF estimates.
According to our experiments, this approach is more efficient and allows to exploit a larger time range than ContinualFlow~\cite{neoral2018continual}.

\begin{figure*}[t]
    \centering
    \includegraphics[width=\linewidth]{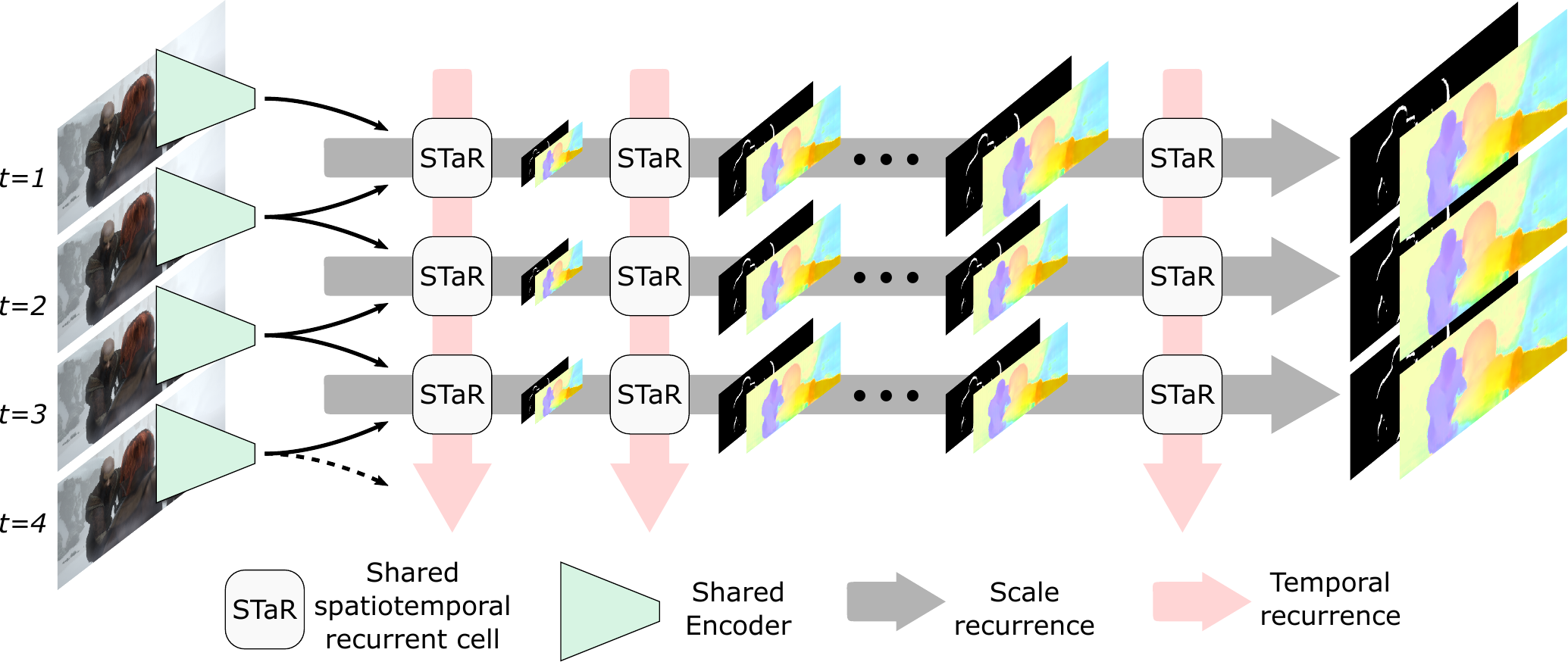}
    \caption{Unrolled view of the proposed SpatioTemporal Recurrent architecture for multi-frame OF estimation (STaRFlow).}
    \label{fig:principles}
\end{figure*}

\subsection{Occlusion Handling}
As OF is ill-defined at occluded pixels, occlusions have to be accounted for during estimation.
Classical methods either treat occlusion as outliers within a robust estimation setting~\cite{brox2004high}, or conduct explicit occlusion detection, often using a forward-backward consistency check \cite{alvarez2007symmetrical}. In a deep learning framework, several methods estimate jointly OF and occlusion maps.
In doing so, most authors (eg. \cite{neoral2018continual,hur2019iterative}) observe a significant improvement on the OF estimation --- an exception being \cite{ilg2018occlusions}.
Unsupervised methods also estimate occlusion maps, as they need to ignore occluded pixels in their photometric loss. \cite{meister2018unflow} estimates occlusion maps by forward-backward
check, \cite{janai2018unsupervised,liu2019selflow} learn occlusion detection in an unsupervised manner.
Very recently, \cite{zhao2020maskflownet} proposes a self-supervised method to learn an occlusion map and uses it to filter the features warping so as to avoid ambiguity due to occlusions.

Here, we propose a very simple and lightweight
way of dealing with occlusions by processing occlusion maps almost in the same way as OF estimates and observe a significant gain on OF accuracy in accordance with \cite{neoral2018continual,hur2019iterative}.

\section{Proposed Approach}
\label{sec:method}

We propose a doubly recurrent algorithm for optical flow (OF) estimation. It is mainly the repeated application of the same \textit{S}patio\textit{T}empor\textit{a}l \textit{R}ecurrent (STaR) cell recursively with respect to time and spatial scale on features extracted from each image of the sequence.
Fig.~\ref{fig:principles} presents an unrolled representation of this recurrent "STaRFlow" model. Feature extraction uses a shared encoder (green block) which architecture comes from~\cite{sun2018pwc}.
The scale recurrence, represented as horizontal gray arrows in Fig.~\ref{fig:principles}, consists in feeding the STaR cell at each scale with the features extracted from the current frame and with the OF and occlusions coming from previous scale. The data flow related to the temporal recurrence carries learned features from one time instant to the next; it is depicted as vertical pink arrows.

The rest of this Section aims at a complete description of STaRFlow.
The internal structure of the STaR cell is presented in section~\ref{sec:method_cell}.
Then section~\ref{sec:method_temporal} focuses on the temporal recurrence, section~\ref{sec:method_occlusions} is dedicated to occlusions handling, and section~\ref{sec:method_spatial} presents the spatial recurrence.
Finally, in Section~\ref{sec:method_loss}, we discuss the compound loss used for multi-frame optical estimation and the optimization process.

\subsection{STaR Cell}
\label{sec:method_cell}
\begin{figure*}[t]
    \centering \includegraphics[width=\linewidth]{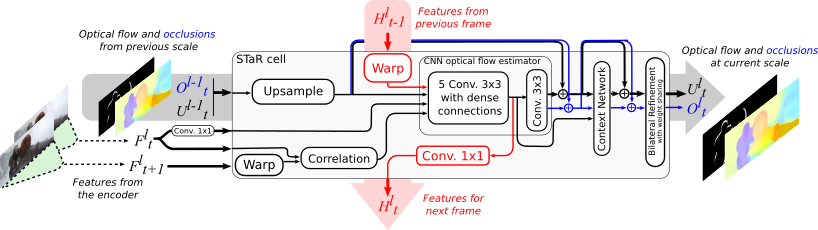}
    \caption{Structure of the proposed SpatioTemporal Recurrent cell (STaR cell).}
    \label{fig:starcell}
\end{figure*}

As several other recent OF estimation approaches, the proposed method builds upon PWC-Net~\cite{sun2018pwc}, which
has been designed to use well-known good practices from energy minimization methods: multi-scale pyramid, warping, cost-volume computation by correlation. These three elements are found in the architecture of the STaR cell presented in Fig.~\ref{fig:starcell}. It is fed by features from a siamese pyramid encoder applied to both frames.
Similarly to PWC-Net, the core trainable block is a CNN dedicated to OF
(blocks \textit{CNN optical flow estimator} and \textit{Context network} in Fig.~\ref{fig:starcell}). Finally, to avoid blurry results
near motion discontinuities,
we use the lightweight bilateral refinement of \cite{hur2019iterative}.

In addition to the inputs already appearing in PWC-Net (features from reference image, cost-volume from correlation of features and the upsampled flow from the previous scale), two supplementary input/output data flows are involved in the STaR cell.
The first one implements the temporal recurrence leading to a multi-frame estimation.
It conveys features from the highest layers of the CNN OF estimator which are fed into the CNN OF estimator at the next time step, see Sec.~\ref{sec:method_temporal}.
The second concerns the occlusion map, which undergoes essentially the same pipeline as the OF --- further details on occlusions handling are given in Sec.~\ref{sec:method_occlusions}.

\subsection{Temporal Recurrence for Multi-Frame Estimation}
\label{sec:method_temporal}
The temporal connection passes features from time $t-1$ to time $t$ (Figure \ref{fig:starcell}). These features are the outputs of the penultimate layer of the CNN OF estimator at $t-1$, which are compressed by a $1 \times 1$ convolution to keep the number of input channels constant from one time step to the next. They are then warped into the current first image geometry, using the previous time-step backward flow \textit{i.e.}, the optical flow from $t$ to $t-1$.
This flow is not directly accessible at inference as our network predicts the forward flow, \textit{i.e.}, from $t-1$ to $t$.
Thus, we apply our network on two frames with reversed time (from $t$ to $t-1$)  to estimate the backward flow (the temporal connection being set to zero).

\subsection{Joint Estimation of Occlusions}
\label{sec:method_occlusions}
As already mentioned, previous works such as \cite{neoral2018continual,hur2019iterative} considered the idea of estimating jointly OF and occlusion maps, with the purpose of improving OF estimation.
In \cite{neoral2018continual} occlusion maps are estimated using an extra CNN module and used as an input of the OF estimator, while \cite{hur2019iterative} processes occlusion map and OF in parallel by adding an occlusion CNN estimator with the same architecture as the OF CNN estimator, but ending with a one-channel sigmoid layer. These methods, especially \cite{hur2019iterative}, lead to a significant increase in the number of parameters of the model.

In the STaR cell, joint estimation of OF and occlusions is done simply by adding a channel to the last convolutional layer of the CNN OF estimator (which, hence, becomes a "OF$+$occlusion" estimator).
After a sigmoid layer, this supplementary channel gives an occlusion probability map with value between 0 (non-occluded) and 1 (occluded).
Compared to \cite{hur2019iterative,neoral2018continual}, this leads to a negligible number of extra parameters, while achieving competitive results, according to the experiments conducted in Sec.~\ref{sec:occlusion_estim}.

\subsection{Spatial Recurrence over Scales}
\label{sec:method_spatial}
We iterate on the same weights on each scale, according to the IRR approach of~\cite{hur2019iterative} --- but unlike them we apply this coarse-to-fine process to a concatenation of the OF and the occlusion map.
This allows a significant decrease in the number of parameters, while keeping estimation results almost unchanged, as shown in Sec.~\ref{sec:ablation:scale}.

\subsection{Multi-Frame Training Loss}
\label{sec:method_loss}
We use $N$-frame training sequences and train our network to estimate the OFs for each pair of consecutive images.
From the second image pair of the sequence, information from previous estimations is transmitted through the temporal connection.
At the end of the sequence, we update the weights so as to decrease:
\begin{equation}
    \mathcal{L} = \frac{1}{N} \sum_{t=1}^N \mathcal{L}_t
\end{equation}
where $\mathcal{L}_t$ is a multi-scale and multi-task loss for image pair ($I_t$, $I_{t+1}$):
\begin{equation}
    \mathcal{L}_t = \sum_{l=1}^L \alpha_l \left(\mathcal{L}_\text{flow}^{t,l} + \lambda \mathcal{L}_\text{occ}^{t,l}\right)
\end{equation}
coefficients $\alpha_l$ being chosen as in \cite{sun2018pwc}.
The supervision of OF $u_t^l(x)$ at each time step $t$ and each scale $l$ is done as in \cite{sun2018pwc} using the $L_2$ norm summed over all pixel positions:
\begin{equation}
    \mathcal{L}_\text{flow}^{t,l} = \sum \left\lVert u_t^l - u_{t,\text{GT}}^l \right\rVert_2
\end{equation}
For the occlusion map $o_t^l$, the loss is a weighted binary cross-entropy:
\begin{equation}
    \mathcal{L}_\text{occ}^{t,l} = - \frac{1}{2} \sum \left( w_t^l o_t^l \log o_{t,\text{GT}}^l + \bar{w}_t^l (1-o_t^l) \log (1-o_{t,\text{GT}}^l) \right)
\end{equation}
where summation is done over all pixel positions and denoting $ w_t^l = \frac{H^l \cdot W^l}{\sum{o_t^l}+\sum{o_{t,\text{GT}}^l}}$ and $\bar{w}_t^l = \frac{H^l \cdot W^l}{\sum(1-o_t^l)+\sum(1-o_{t,\text{GT}}^l)}$, $H^l$ and $W^l$ being the image size at scale $l$. As in \cite{hur2019iterative} we update at each iteration the weight $\lambda$ that balances the flow loss and the occlusion loss.

\section{Experiments} \label{sec:experiments}

\subsection{Implementation Details}\label{sec:details}

As proposed in \cite{ilg2017flownet2}, all models are first trained on FlyingChairs \cite{dosovitskiy2015flownet} and then on FlyingThings3D \cite{mayer2016large}. We then finetune on either Kitti or MPI Sintel.
We use photometric and geometric data augmentations as in \cite{hur2019iterative} except that for the geometric augmentations we do not apply relative transformations.

\subsubsection{Pretraining on Image Pairs on FlyingChairs}
Following \cite{neoral2018continual}, we first train our multi-frame architecture, except from the temporal connection, on 2D two-frame data. To supervise both OF and occlusion estimation, we use the FlyingChairsOcc dataset \cite{hur2019iterative}.
We train with a batch size of 8 for 600k iterations, with an initial learning rate of $10^{-4}$ which is divided by 2 every 100k iterations after the first 300k iterations.

\subsubsection{Multi-Frame Training on FlyingThings3D}
Then we train the STaRFlow model on sequences of $N=4$ images from FlyingThings3D, the temporal data stream being initialized to zero --- note that longer sequences could be exploited,
at the cost of an increase in the memory space required for training.
As it is the first training for the temporal connection, we start with a higher learning rate of $10^{-4}$ compared to two-frame training (as suggested by \cite{neoral2018continual}) and train for 400k iterations, dividing the learning rate by 2 every 100k iterations after the first 150k iterations. We use a batch size of 4.
For the ablation study, this is the final step of our training.

\subsubsection{Finetuning on MPI Sintel or Kitti}

We use the same finetuning protocol as \cite{hur2019iterative} but extended to our multi-frame ($N=4$) estimation process.
For Sintel, we can supervise every time step. In KITTI, only one time step is annotated, hence we only supervise the last time-step estimation.
This finetuning step is only used for benchmark submissions.

\subsubsection{Running Time}
On Sintel images ($1024 \times 436$) the inference time of STaRFlow is of 0.22 second per image pair, on a mid-range NVIDIA GTX 1070 GPU.

\subsection{Optical Flow Results on Benchmarks}\label{sec:bench}

Results of STaRFlow on benchmarks MPI Sintel and KITTI 2015 are given in Tab.~\ref{tab:benchmarks_results}, and compared to top-leading methods and/or methods closely related to our approach.
STaRFlow reaches the best EPE score on the final pass of Sintel, is second on the clean pass, and is on par with IRR-PWC on Kitti2015. Kitti2015 is characterized by very large movements of foreground objects, which generally disadvantages multi-frame methods: among them, STaRFlow still ranks second behind MFF. Regarding the number of parameters, STaRFlow ranks second behind ARFlow but outperforms it (as well as other light methods such as LiteFLowNet and SelFlow) in terms of OF precision. It is also interesting to compare STaRFlow with the related methods \cite{neoral2018continual} and \cite{hur2019iterative}. STaRFlow significantly outperforms ContinualFlow~\cite{neoral2018continual} on all benchmarks while being three times lighter. Compared to IRR-PWC \cite{hur2019iterative}, the benefit of the multi-frame estimation of STaRFlow clearly appears on MPI Sintel.

\begin{table}
    \centering
    \caption{Results on MPI Sintel and KITTI 2015 benchmarks (test sets). Endpoint error [px] on Sintel, percentage of outliers on KITTI. }
    \begin{tabular}{|l|cc|c|c|}
        \hline
        \multirow{2}{*}{Method} & \multicolumn{2}{c|}{MPI Sintel} & KITTI 2015 & Number of\\
         & clean & final & Fl-all & parameters \\
         \hline
         ARFlow-mv$^*$ \cite{liu2020arflow} & 4.49 & 5.67 & 11.79 \% & \textbf{2.37M} \\
         LiteFlowNet \cite{hui18liteflownet} & 4.54 & 5.38 & 9.38 \% & 5.37M \\
         PWC-Net \cite{sun2018pwc} & 4.39 & 5.04 & 9.60 \% & 8.75M  \\
         LiteFlowNet2 \cite{hui2019lightweight} & 3.48 & 4.69 & 7.62 \% & 6.42M \\
         PWC-Net+ \cite{sun2018models} & 3.45 & 4.60 & 7.72 \% & 8.75M$^{\dag}$ \\
         IRR-PWC \cite{hur2019iterative} & 3.84 & 4.58 & 7.65 \% & 6.36M \\
         MFF$^*$ \cite{ren2019fusion} & 3.42 & 4.57 & 7.17 \% & 9.95M \\
         ContinualFlow\_ROB$^*$ \cite{neoral2018continual} & 3.34 & 4.53 & 10.03 \% & 14.6M$^{\dag}$ \\
         SelFlow$^*$ \cite{liu2019selflow} & 3.74 & 4.26 & 8.42 \% & 4.79M$^{\ddag}$ \\
         MaskFlowNet \cite{zhao2020maskflownet} & \textbf{2.52} & 4.17 & \textbf{6.11 \%} & N/A \\
         ScopeFlow \cite{bar2020scopeflow} & 3.59 & \textit{4.10} &  \textit{6.82} \% & 6.36M \\
         \hline
         STaRFlow-ft$^*$ (\emph{ours}) & \textit{2.72} & \textbf{3.71} & 7.65 \% & \textit{4.77M} \\
         \hline
    \end{tabular}
    \begin{center}
    Best results are in bold characters, second ones in italic. Multi-frame methods are marked with $^*$.  $\dag$: value given in \cite{hur2019iterative}, $\ddag$: value given in \cite{liu2020arflow}.  \\
    \end{center}

    \label{tab:benchmarks_results}
\end{table}

\subsection{Occlusion Estimation}\label{sec:occlusion_estim}

Our main purpose here is to compare our solution for occlusion estimation, which shares almost all its weights with the OF estimator, to the dedicated decoder used in IRR-PWC. To do this comparison as fairly as possible, we have trained a two-frame version of STaRFlow (by removing the red connections and operators on Fig.~\ref{fig:starcell}), which then essentially differs from IRR-PWC by the occlusion detection process.
In Tab.~\ref{tab:estim_occ}, we compare F1-scores of our occlusion estimator and various methods (including IRR-PWC) on occlusion maps estimated from MPI Sintel data.
Our occlusion estimation is on par with IRR-PWC while being much lighter.
We also report scores of SelFlow and ScopeFlow for comparison to other state-of-the-art methods.

\begin{table}
    \centering
    \caption{Occlusion map estimation results (F1-score) on MPI Sintel.}
    \begin{tabular}{|l|cc|c|}
        \hline
        Method & Clean & Final & Parameters\\
        \hline
        ContinualFlow \cite{neoral2018continual} & - & 0.48 & 14.6M \\
        SelFlow \cite{liu2019selflow} & 0.59 & 0.52 & 4.79M \\
        IRR-PWC \cite{hur2019iterative} & 0.71 & 0.67 & 6.36M \\
        ScopeFlow \cite{bar2020scopeflow} & \textbf{0.74} & \textbf{0.71} & 6.36M \\
        \hline
        Our occlusion estimator & 0.70 & 0.66 & \textbf{4.09M} \\
        \hline
    \end{tabular}
    \begin{center}
    Best results are in bold characters.\\
    \end{center}
    \label{tab:estim_occ}
\end{table}

\begin{table*}
    \centering
    \caption{Influence of temporal connection and occlusion modules on performances (MPI Sintel and KITTI 2015 training sets).}
    \begin{tabular}{|ll|ccc|ccc|cc||cc|}
        \hline
        \multirow{2}{*}{Method}& Cat. & \multicolumn{3}{c}{Sintel Clean [px]} & \multicolumn{3}{|c}{Sintel Final [px]} &  \multicolumn{2}{|c||}{KITTI 2015} & \multicolumn{2}{c|}{Parameters}\\
         & &all & noc & occ & all & noc & occ & epe-all & Fl-all & number & relative \\
         \hline
         \multicolumn{12}{|l|}{\textit{Without joint occlusion estimation.}} \\
         \hline
         Backbone (PWC-Net)         & 2F    & 2.74 &  1.46 & 16.48  & 4.18 &  2.56 & 21.70 & 11.75 & 33.20 \% & 8.64M & 0 \% \\
         Backbone + TRFlow      & MF  & 2.47 &  \textbf{1.41} & 13.97  & 4.01 &  2.52 & 20.00 & 11.27 & 33.77 \% & 8.68M & $+$0.5 \% \\
         Backbone + TRFeat   &  MF  & \textbf{2.45} & 1.44 & \textbf{13.36} & \textbf{3.76} & \textbf{2.46} & \textbf{17.82} & \textbf{9.94} & \textbf{32.12 \%} & 12.31M & $+$42.5 \% \\
         \hline
         \hline
         \multicolumn{12}{|l|}{\textit{With joint occlusion estimation.}} \\
         \hline
         Backbone     &  2F    & 2.46 &  1.32 & 14.82  & 3.96 &  2.47 & 20.06 & 10.58 & 31.28 \% & 8.68M & $+$0.5 \% \\
         Backbone + TRFlow  & MF   & 2.17 &  1.23 & 12.33  & 3.90 &  2.50 & 19.11 & 10.82 & 32.51 \% & 8.73M & $+$1.0 \% \\
         Backbone + TRFeat & MF   &\textbf{2.09} & \textbf{1.21} & \textbf{11.63} & \textbf{3.43} & \textbf{2.24} & \textbf{16.24} & \textbf{8.79} & \textbf{28.18 \%} & 12.38M & $+$43.3 \% \\
         \hline
         \hline
         \multicolumn{12}{|l|}{\textit{With joint occlusion estimation and spatial recurrence.}} \\
         \hline
         Backbone     &  2F    & 2.29 &  \textbf{1.20} & 14.03 & 3.72 &  \textbf{2.32} & 18.77 & 10.74 & 31.35 \% & 3.37M & $-$61.0 \% \\
         Backbone + TRFlow  & MF   & 2.20 &  1.25 & 12.40  & 3.98 &  2.56 & 19.38 & 11.00 & 35.23 \% & 3.38M & $-$60.9 \% \\
         Backbone + TRFeat & MF   & \textbf{2.10} &  1.22 & \textbf{11.67} & \textbf{3.49} &  \textbf{2.32} & \textbf{16.15} & \textbf{9.26} & \textbf{30.75} \% & 4.37M & $-$49.4 \% \\
         \hline
    \end{tabular}

    \vspace{0.1cm}
    Best results are in bold characters. Fl-all, on KITTI, is the percentage of outliers (epe $>3$ px).\\
    2F (resp. MF) refers to two-frame (resp. multi-frame) methods. \textit{TR} stands for \textit{temporal recurrence}.

    \label{tab:abl_temp}
\end{table*}

\subsection{Ablation Study}\label{sec:ablation}

In this section, we consider the contributions of the following components of the STaRFlow model to the OF estimation: temporal recurrence and number of used frames, joint occlusion estimation and spatial recurrence.
For all the experiments, our backbone is the two-frame PWC-Net architecture \cite{sun2018pwc}\footnote{Implementation from \url{https://github.com/visinf/irr}}
that we trained as described in \cite{hur2019iterative}.
As this backbone does not include a bilateral refinement module, we do not include this module in the following tests.
The models are trained on FlyingChairs and FlyingThings3D, without any further finetuning, and tested on the training sets of MPI Sintel and KITTI2015.
All comparisons are made with the main performance metrics proposed in the benchmark websites --- note that we use the revised occlusion maps provided by \cite{hur2019iterative} to compute occ/noc scores on MPI Sintel.

\subsubsection{Temporal Recurrence}\label{sec:ablation:temporal}

Two different temporal recurrences are evaluated, with and without occlusion handling in Tab.~\ref{tab:abl_temp}, and compared to the two-frame backbone. The first one, termed "TRFlow", is inspired from \cite{neoral2018continual}, and passes the estimated OF at time $t-1$ to the CNN OF estimator at $t$. In the second approach, denoted by "TRFeat", the temporal connection conveys learned features. "TRFeat" is the method implemented in STaRFlow and described in Sec.~\ref{sec:method_temporal}.

According to Tab.~\ref{tab:abl_temp}, using learned features in the temporal connection yields better results than passing estimated OFs, with higher EPE gains on degraded images (Sintel Final vs. Sintel Clean) and especially on the real images of KITTI2015 training dataset. Results are consistent whether an occlusion module is used or not.

\begin{figure*}[t]
\begin{center}
    \includegraphics[width=1\linewidth]{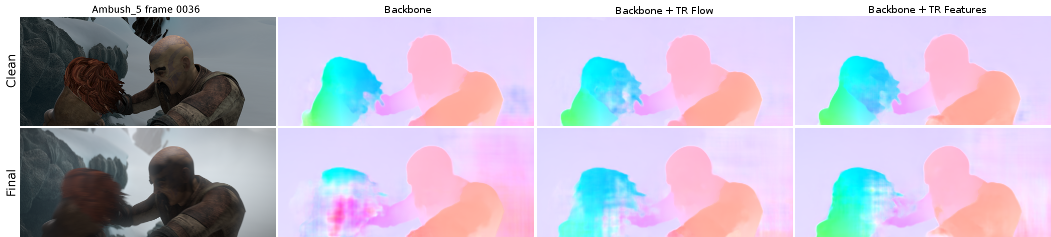}
\end{center}
\caption{Multi-frame estimation provides robustness to degraded image quality: results on Sintel clean (upper row) and Sintel Final pass (lower row).}
\label{fig:degradedimg}
\end{figure*}

\begin{figure*}[t]
\begin{center}
    \includegraphics[width=1\linewidth]{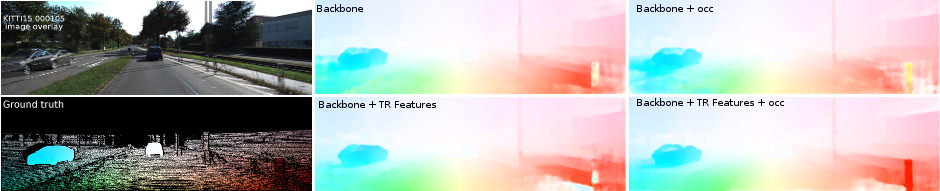}
\end{center}
\caption{Both the occlusion and temporal coherence modules are needed here to resolve the motion of the lower right roadsign.}
\label{fig:needsocc}
\end{figure*}

\begin{figure*}[t]
\begin{center}
    \includegraphics[width=1\linewidth]{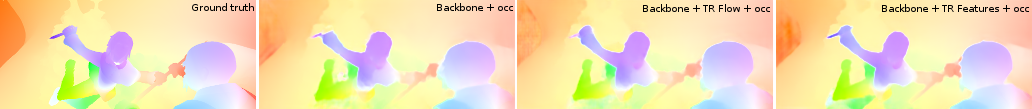}
\end{center}
\caption{Our temporal recurrent cell improves optical flow estimation of small objects.}
\label{fig:smallobject}
\end{figure*}

The qualitative results displayed in Fig.~\ref{fig:degradedimg}--\ref{fig:smallobject} aim to better understand the gains brought by our temporal connection and occlusions handling. As could be expected, multi-frame estimation improves robustness to degraded image quality. This is shown in Fig.~\ref{fig:degradedimg} which compares results on Sintel Clean and Final (blurry) images.

Multi-frame estimation also allows temporal inpainting: for a region occluded at time $t+1$ but visible at $t$ and previous time steps, the previously estimated motion can be used to predict the motion between $t$ and $t+1$. This could be observed on the Sintel example shown in the upper left part of Fig.~\ref{fig:test_final}: the right knee of the central character, although occluded in the next frame, is correctly estimated by STaRFlow. Fig.~\ref{fig:needsocc} displays an example extracted from KITTI2015 training set where temporal connection and occlusions estimation are both required to correctly estimate motion of the roadsign on the lower right part of the image, which is occluded in the next frame. Finally, Fig.~\ref{fig:smallobject} shows that our temporal connection with learned features yields increased sensitivity to small object motion compared to the backbone and also to TRFlow.

\subsubsection{Occlusion Handling}\label{sec:ablation:occlusion}

Comparison of methods with and without occlusion estimation in Tab.~\ref{tab:abl_temp} shows that adding the task of detecting occlusions consistently helps OF estimation. This is true for two-frame and multi-frame methods.

\subsubsection{Spatial Recurrence}\label{sec:ablation:scale}

The lower part of Tab.~\ref{tab:abl_temp} is devoted to the spatial recurrence, \textit{i.e.} the iterations on the same weights over scales in the coarse-to-fine multi-level estimation \cite{hur2019iterative}. While OF precision is only marginally affected by this implementation, large gains in terms of number of parameters are obtained with respect to the PWC-Net backbone (see last column).

\begin{figure*}[t]
\begin{center}
    \includegraphics[width=1\linewidth]{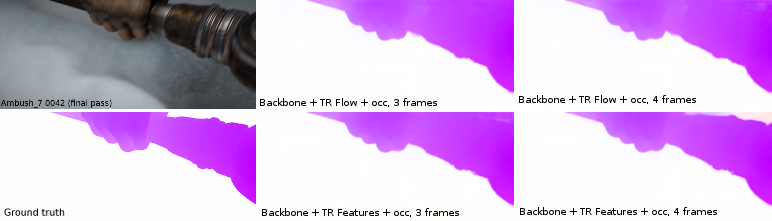}
\end{center}
\caption{The benefit of exploiting more frames in OF estimation for sequence Ambush7 of Sintel Final.}
\label{fig:nframes}
\end{figure*}

\subsubsection{Impact of the Number of Frames at Test Time}

Recall that $N=4$ frames are used for training multi-frame models (TRFlow and TRFeat). It means that, at training time, the temporal connection is reinitialized to zero every 4 frames, essentially to avoid an increased memory cost, beyond the capacity of the hardware. However, \emph{at test time}, the temporal connection can be exploited  over a different time horizon. This is the object of Tab.~\ref{tab:abl_nframes}, which compares temporal connections TRFlow and TRFeat when increasing the number of frames $N'$ used at test time. Each line of the Table presents scores computed for the OF estimated between time instants $N'-1$ and $N'$.

According to Tab.~\ref{tab:abl_nframes}, performance improves more for TRFeat than for TRFlow when $N'$ increases. This is particularly true for degraded (Sintel Final) or real images (KITTI), or in occluded regions. Furthermore, we observe that TRFeat still improves using $N'=5$ frames. TRFeat, by propagating learned features in the temporal connection instead of OF, exploits more efficiently long term memory than TRFlow and appears even able to learn a temporal continuity beyond the number of frames used for training.

This can also be seen on the qualitative results presented on Fig.~\ref{fig:nframes}. Estimations using $N'=3$ and $N'=4$ (columns 2 and 3) are presented for TRFlow and TRFeat. The fact that the object is very close to the image border makes the problem difficult. For the two methods, using 3 frames is not enough to correctly estimate the object's contour. TRFeat manage to resolve the contour with a 4th frame, while TRFlow still fails to do it.

\begin{table}
    \centering
    \caption{Impact of the number of frames $N'$ used at test time.}
    \begin{tabular}{|c|c@{\hskip 1em}c@{\hskip 1em}c|c@{\hskip 1em}c@{\hskip 1em}c|c@{\hskip 1em}c|}
                \multicolumn{9}{c}{Backbone + occ + TRFlow + SR} \\
                \hline
                & \multicolumn{3}{c}{Sintel Clean} & \multicolumn{3}{|c}{Sintel Final} &  \multicolumn{2}{|c|}{Kitti15} \\
                $N'$ & all & noc & occ & all & noc & occ & epe-all & Fl-all \\
        \hline
                2 & 2.36 &  1.27 & 14.17  & 4.05 &  2.57 & 20.06 & 12.53 & 35.95 \% \\
                3 & \textbf{2.17} & \textbf{1.24} & \textbf{12.29}  & \textbf{3.95} & \textbf{2.56} & \textbf{19.03} & 11.26 & 35.35 \% \\
                4 & 2.20 &  1.25 & 12.40  & 3.98 & \textbf{2.56} & 19.38 & 11.01 & 35.27 \% \\
                5 & 2.20 &  1.26 & 12.37  & 3.98 & \textbf{2.56} & 19.30 & \textbf{10.94} & \textbf{35.17 \%} \\
                6 & 2.20 &  1.26 & 12.33  & 3.98 &  2.58 & 19.11 & \textbf{10.94} & 35.19 \% \\
                \hline
                \multicolumn{9}{c}{} \\
                \multicolumn{9}{c}{Backbone + occ + TRFeat + SR} \\
                \hline
                & \multicolumn{3}{c}{Sintel Clean} & \multicolumn{3}{|c}{Sintel Final} &  \multicolumn{2}{|c|}{Kitti15} \\
                $N'$ & all & noc & occ & all & noc & occ & epe-all & Fl-all \\
        \hline
                2 & 2.40 &  1.30 & 14.34  & 4.04 &  2.55 & 20.12 & 12.01 & 34.22 \% \\
                3 & 2.10 &  1.23 & 11.60  & 3.58 &  2.35 & 16.90 & 9.95 & 31.49 \% \\
                4 & 2.10 & \textbf{1.22} & 11.67  & 3.49 &  2.32 & 16.15 & 9.26 & 30.78 \% \\
                5 & \textbf{2.08} & \textbf{1.22} & \textbf{11.36}  & \textbf{3.43} & \textbf{2.27} & \textbf{15.99} & 9.17 & \textbf{30.66 \%} \\
                6 & 2.09 & \textbf{1.22} & 11.52  & 3.50 &  2.32 & 16.25 & \textbf{9.14} & 30.69 \% \\
        \hline
    \end{tabular}
    \begin{center}
    Best results are in bold characters.\\
    \end{center}
    \label{tab:abl_nframes}
\end{table}

\section{Conclusion}\label{sec:conclusion}

We have presented STaRFlow, a new lightweight CNN method for multi-frame OF estimation with occlusion handling. It involves a unique computing cell which recursively processes both a spatial data flow in a coarse-to-fine multi-scale scheme and a temporal flow which conveys learned features. Using learned features in the temporal recurrence allows better exploitation of temporal information than propagating OF estimates as proposed in \cite{neoral2018continual}. STaRFlow builds upon approaches such as~\cite{hu2018recurrent,hur2019iterative} based on the repeated use of the same weights over a scale recurrence but extends this idea to a double time-scale recurrence. Moreover, we have also shown that occlusion estimation can be done with a minimal number of extra parameters, simply by adding a dedicated layer to the output tensor of the CNN OF estimator.
STaRFlow gives state-of-the-art results on the two benchmarks MPI Sintel and Kitti2015, even outperforming, at the time of writing, all previously published methods on Sintel final pass. Moreover, STaRFlow is lighter than all other two-frame or multi-frame methods with comparable performance.

Quantitative and qualitative evaluations on MPI Sintel and KITTI2015 show that STaRFlow improves OF quality on degraded images and on small objects thanks to temporal redundancy, and is also able to achieve efficient temporal inpainting in occluded areas. Our experiments also confirm conclusions of~\cite{hur2019iterative,neoral2018continual} that learning to predict occlusions consistently improves OF estimation. Moreover, our implementation, based on sharing almost all weights between OF and occlusion estimation, further indicates that these two tasks are closely related one to the other.

\section*{Acknowledgments}
The authors are grateful to the French agency of defense (DGA) for financial support, and to the ONERA project DELTA.

\bibliographystyle{plain}
\bibliography{ourbib}

\end{document}